\documentclass[11pt]{article}

\usepackage[]{acl}

\usepackage{times}
\usepackage{latexsym}
\usepackage{multirow}

\usepackage[T1]{fontenc}

\usepackage[utf8]{inputenc}
\usepackage{CJKutf8}

\usepackage{microtype}

\usepackage{amsmath}
\usepackage{amssymb}
\usepackage{graphicx}
\usepackage{url}
\usepackage{subfigure}

\title{Extending Word-Level Quality Estimation for Post-Editing Assistance}

\author{
  {\bf Yizhen Wei}$^\dagger$ \
  {\bf Takehito Utsuro}$^\dagger$ \
  {\bf Masaaki Nagata}$^\ddagger$ \\
$^\dagger${Deg. Prog. Sys.\&Inf. Eng., Grad. Sch. Sci.\&Tech., University of Tsukuba}
\\
$^{\ddagger}${NTT Communication Science Laboratories, NTT Corporation, Japan}
}

\begin{document}
\maketitle

\begin{abstract}
We define a novel concept called extended word alignment in order to improve post-editing assistance efficiency.
Based on extended word alignment, we further propose a novel task called refined word-level QE that outputs refined tags and word-level correspondences.
Compared to original word-level QE, the new task is able to directly point out editing operations, thus improves efficiency.
To extract extended word alignment, we adopt a supervised method based on mBERT. 
To solve refined word-level QE, we firstly predict original QE tags by training a regression model for sequence tagging based on mBERT and XLM-R.
Then, we refine original word tags with extended word alignment.
In addition, we extract source-gap correspondences, meanwhile, obtaining gap tags.
Experiments on two language pairs show the feasibility of our method and give us inspirations for further improvement.
\end{abstract}

\section{Introduction}
Post-editing refers to the process of editing a rough machine-translated sentence (referred to as MT) into a correct one.
Compared with conventional statistical machine translation \citep{koehn-etal-2003-statistical}, neural machine translation \citep{cho-etal-2014-learning, sutskever-2014-sequence, vaswani-2017-attention} can generate translations with high accuracy.
However, \citet{yamada-2019-impact} suggested that there is no significant difference in terms of cognitive load for one to post-edit an MT even it has high quality.
Therefore, post-editing assistance is profoundly needed.

Traditional post-editing assistance methods leave room for improvement.
A typical method is word-level QE \citep{WMT20} that predicts tags expressed in the form of \textbf{OK} or \textbf{BAD}.
However, such a dualistic judgement is not efficient enough because meaning of \textbf{BAD} is ambiguous.
\begin{figure*}
\centering
\subfigure[Original word-level QE]{
\includegraphics[width=75mm]{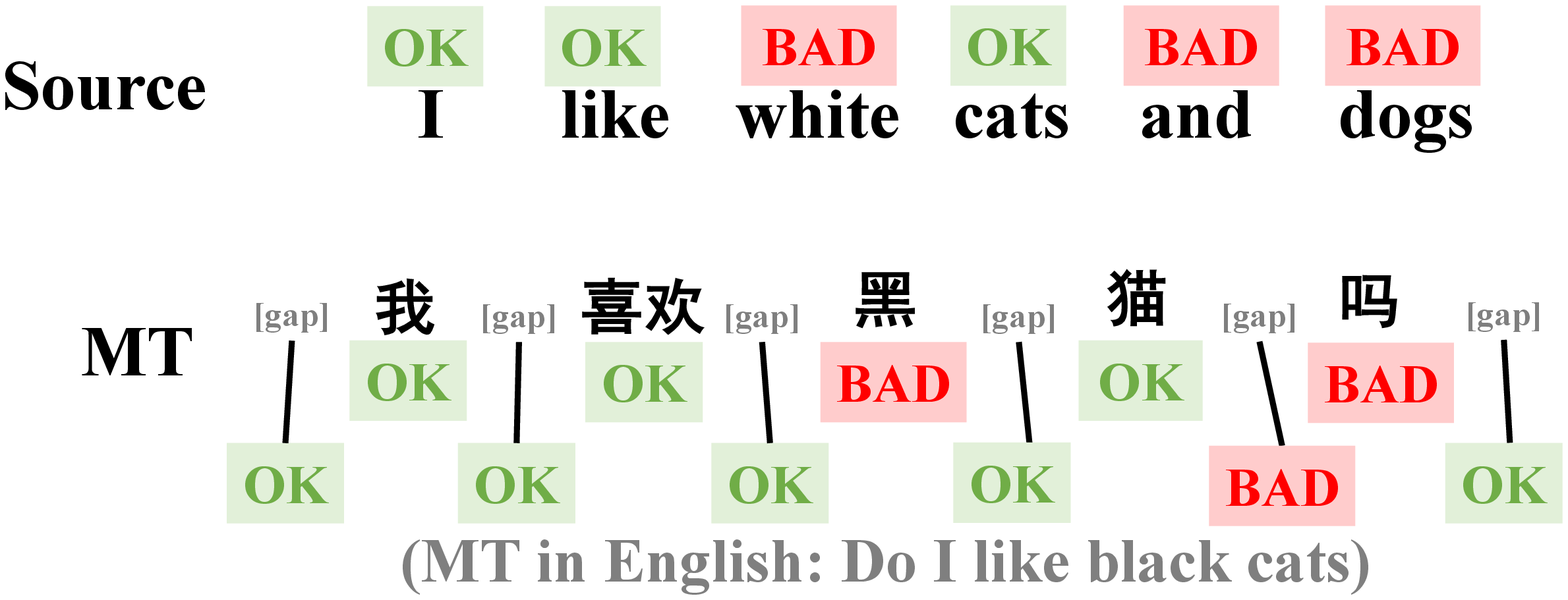}
\label{fig:demo_noalign}
}
\subfigure[Refined word-level QE. Correspondences between \textbf{REP} are drawn in red and that between \textbf{INS} are drawn in purple.]{
\includegraphics[width=75mm]{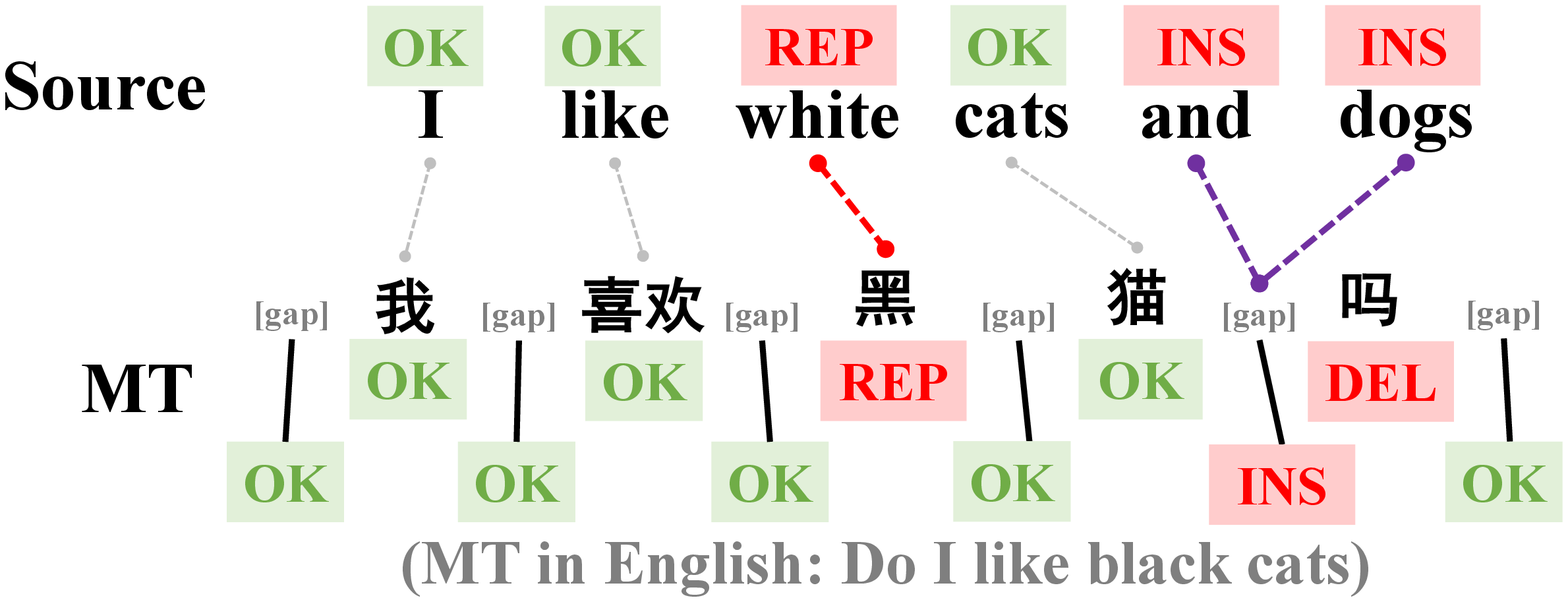}
\label{fig:demo_align}
}
\caption{A comparison between original word-level QE and our proposal.}
\label{fig:demo}
\end{figure*}

Word alignment is also proved to be helpful for post-editing assistance.
\citet{schwartz-etal-2015-effects} demonstrated that displaying word alignment statistically significantly improves post-editing quality.
However, unlike QE tags, word alignment cannot tell where translation errors are.
Besides, it is non-trivial to extract word alignment between source sentence and MT.
\citet{schwartz-etal-2015-effects} used a built-in function of Moses \citep{koehn-etal-2007-moses}, a decoder for statistical machine translation that is no longer suitable for neural models.

In this paper, we propose a novel concept called extended word alignment.
In extended word alignment, we include incorrect word translations and null alignment between a source sentence and MT.
We adopt a supervised method based on pre-trained language models to extract it.
Based on extended word alignment, we further propose a novel task called refined word-level QE which outputs refined tags including \textbf{REP}, \textbf{INS},  and \textbf{DEL} along with word-level correspondences.
By referring to those information, post-editors could immediately realize what operations (replacement, insertion, and deletion towards MT) to do.
Thus, we believe that refined word-level QE can significantly improve post-editing assistance efficiency.
Methodologically, we firstly predict original word tags by training regression models for sequence tagging based on architectures such as multilingual BERT \citep{delvin-2019-bert} (mBERT) and XLM-RoBERTa \citep{conneau-2020-unsupervised} (XLM-R).
Then, we refine the original word tags by incorporating extended word alignment in a rule-based manner.
In addition, we adopt a method similar to the one for extended word alignment to extract source-gap correspondences and then determine gap tags.


Experiments on En-De and En-Zh datasets are conducted.
Results show that our method significantly outperforms the baseline.
For En-De, our best performance outperforms the baseline by 12.9\% and 6.0\% respectively in terms of mean F1 scores for Source and MT word refined tags.
For En-Zh, the gap reaches 48.9\% and 16.9\%.
Further more, we discuss the effectiveness and limitations of our method with specific cases.

\section{Related Work}

\textbf{Word Alignment Extraction.}
Methods based on statistical models \citep{brown-etal-1993-mathematics, och-ney-2003-systematic, dyer-etal-2013-simple} were dominant methods for word alignment extraction.
In recent years, neural-based methods developed quickly.
\citet{garg-2019-jointly} tried to obtain word alignment based on attention inside a transformer \citep{vaswani-2017-attention}, but their method perform just as well as statistical tools like GIZA++ \citep{och-ney-2003-systematic}.
\citet{dou-2021-word} utilized multilingual BERT to extract embeddings of all words conditioned on context, aligning them under the restriction of optimal transport \citep{kusner-2015-word}.
\citet{nagata-etal-2020-supervised} utilized the pre-trained language model in a supervised manner and achieved a significant improvement against previous studies with only around 300 parallel sentence pairs for fine-tuning.
In our work, we adapt their approach from ordinary word alignment to extended word alignment.
Details will be introduced in Section \ref{sec:align by mbert}.

\textbf{Word-Level QE.}
One of the conventional architectures for word-level QE is LTSM-based predictor-estimator \citep{kim-lee-2016-recurrent,zhang-weiss-2016-stack,kim-2017-predictor}.
Recent researches \citep{wang-EtAl:2020:WMT4} adopted new architectures such as transformer \citep{vaswani-2017-attention}.

For moderner methods, a typical example is QE BERT \citep{kim-2019-qebert}.
They built a mBERT for classification with explicit gap tokens in the input sequence, but we find that regression models with adjustable threshold consistently outperform classification models and explicit gap tokens harm final performance.
A newer research \citep{lee-2020-two} adopted XLM-R rather than mBERT, but they did not explain their strategy to determine a threshold.

All methods above require third-party large-scale parallel data for pre-training.
In contrast, our method introduced in Section \ref{sec:original tag prediction} achieves acceptable performance with small cost.

\textbf{Post-Editing User Interface.}
\citet{nayek-etal-2015-catalog} depicted an interface where words that need editing are displayed with different colors. 
\citet{schwartz-etal-2015-effects} emphasized the importance of displaying the word alignment.
Both interfaces do not tell the correctness of the translation of the MT words.
Compared to them, the interface we envisaged provides information about translation quality (correctness) as well as suggestions of specific post-editing operations.

There are also some other studies\citep{herbig-etal-2020-mmpe,albo-jamara-etal-2021-mid} tried to introduce multi-modalities including touching, speech, hand gestures into post-editing user interface, improving efficiency from another perspective.

\section{Refined Word-Level QE for Post-Editing Assistance}

\subsection{Original Word-Level QE}
\label{sec:original word-level qe}
According to \citet{WMT20}, word-level QE shown in Figure \ref{fig:demo_noalign} is a task that takes a source sentence and its machine-translated counterpart (MT) as input.
It then outputs tags for source words, MT words and gaps between MT words (MT gaps).\footnote{For convenience, source tags and MT word tags are collectively known as word tags. MT word tags and MT gap tags are collectively known as MT tags.}
All those tags are expressed either as \textbf{OK} or \textbf{BAD}.
\textbf{BAD} indicates potential translation errors that post-editors should correct.
We refer to such a task as original word-level QE.

Original word-level QE is not efficient enough for post-editing assistance because \textbf{BAD} is ambiguous.
For example, in Figure \ref{fig:demo_noalign}, tag of ``white'' indicates a replacement of the mistranslation ``\begin{CJK}{UTF8}{gbsn}黑\end{CJK}'' (black), but tag of ``dogs'' indicates an insertion into the gap between ``\begin{CJK}{UTF8}{gbsn}猫\end{CJK}'' and ``\begin{CJK}{UTF8}{gbsn}吗\end{CJK}''.
It is impossible to distinguish between these indications unless one attend to both entire sentences, which makes post-editing assistance meaningless.

\subsection{Extended Word Alignment}
We formally define a novel concept called extended word alignment between source sentence and MT.
Ordinary word alignment indicates word-to-word relations between a pair of semantically equivalent sentences in two languages.
Any word can be theoretically aligned with another semantically equivalent word on the other side.
In contrast, in extended word alignment, translation errors in MT are considered.
Specifically, a source word is allowed to be aligned with its mistranslation (wrong word choice) and a word is allowed to be aligned with nothing, namely null-aligned.


\subsection{Refined Word-Level QE}
\label{subsec:refined_qe_tags}
Extended word alignment can disambiguate \textbf{BAD} tags, overcoming the disadvantage of original word-level QE.
When a \textbf{BAD}-tagged source word is aligned with a \textbf{BAD}-tagged MT word, it is clear that a replacement is needed.
Likewise, a null-aligned \textbf{BAD}-tagged source word indicates an insertion and a \textbf{BAD}-tagged MT word is a deletion.

To make our idea more user-friendly, we formally propose a novel task called refined word-level QE by incorporating extended word alignment with original word-level QE.
Besides extended word alignment, following refined tags are also included as the objectives.
\begin{itemize}
\setlength{\itemsep}{0pt}
\setlength{\parsep}{0pt}
\setlength{\parskip}{0pt}
\item \textbf{REP} is assigned to a source word and its mistranslation (wrong word choice) in MT, indicating a replacement.
\item \textbf{INS} is assigned to a source word and the gap where translation should be inserted in, indicating an insertion.
\item \textbf{DEL} is assigned to a redundant MT word, indicating a deletion.
\end{itemize}
In addition, we include correspondences between \textbf{INS}-tagged source words and MT gaps to express the insertion points.
Those source-gap correspondences along with extended word alignment are collectively referred to as word-level correspondences.

Figure \ref{fig:demo_align} is an example of our proposal.
Compared with Figure \ref{fig:demo_noalign}, Figure \ref{fig:demo_align} points out the replacement of ``\begin{CJK}{UTF8}{gbsn}黑\end{CJK}'' (black), the insertions of ``and'' and ``dogs'' to the insertion point, and the deletion of ``\begin{CJK}{UTF8}{gbsn}吗\end{CJK}'' (an interrogative voice auxiliary).

\section{Methodology\footnote{Besides the current method, we have also tried to use a unified model based on architectures like XLM-R to directly predict refined tags (\textbf{OK}/\textbf{REP}/\textbf{INS}/\textbf{DEL}) and word-level correspondences. However, due to lack of training data and complexity of the problem, direct approach did not work well. Therefore, we decided to adopt this multiple-phase approach.}}
\label{sec:methodology}

\subsection{Extended Word Alignment Extraction}
\label{sec:align by mbert}
\begin{figure}
\centering
\includegraphics[width=75mm]{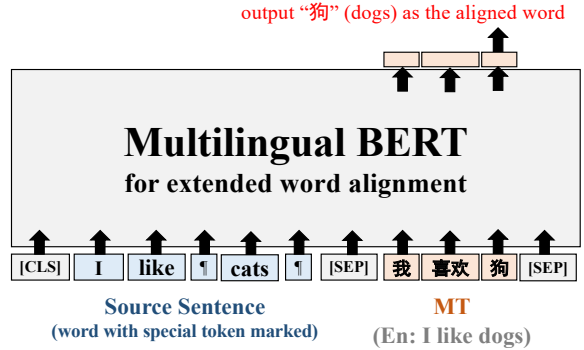}
\caption{Extracting extended word alignment by mBERT. The word will be aligned with [CLS] token if it is null-aligned.
}
\label{fig:align by mbert}
\end{figure}
Extracting extended word alignment is non-trivial.
Traditional unsupervised statistical tools\citep{och-ney-2003-systematic,dyer-etal-2013-simple} cannot work well because they expect semantically equivalent sentence pair as input.
After trying several neural methods \citep{garg-2019-jointly, dou-2021-word}, we empirically adopt the supervised method proposed by \citet{nagata-etal-2020-supervised}.

Specifically, extended word alignment extraction is regarded as a cross-lingual span prediction problem similar to the paradigm that utilizes BERT \citep{delvin-2019-bert} for SQuAD v2.0 \citep{rajpurkar-etal-2018-know}.
mBERT is used as the basic architecture.
Given a source sentence with one word marked $S=[s_1,s_2,...,M, s_i, M,...,s_m]$ ($M$ stands for a special mark token) and the MT $T=[t_1,t_2,...,t_n]$, mBERT is trained to identify a span $T_{(j,k)}=[t_j,...,t_k] (1 \leq j \leq k \leq n)$ that is aligned with the marked source word $s_i$.
Cross entropy loss is adopted during training.
\begin{center}
$\mathcal{L}^{align}_{s_i} = -\frac{1}{2}[\mathrm{log}(p_{j}^{start}) + \mathrm{log}(p_k^{end})]$	
\end{center}
Because of the symmetry of word alignment, similar operations will be done again in the opposite direction.
During testing, following \citet{nagata-etal-2020-supervised}, we recognize word pairs whose mean probability of both directions is greater than 0.4 as a valid word alignment.
The image of the model is illustrated in Figure \ref{fig:align by mbert}.

\citet{nagata-etal-2020-supervised} demonstrated that the mBERT-based method significantly outperforms statistical methods in ordinary word alignment extraction.
According to them, extracting word alignment for each word independently is the key to outperform other methods.
Traditional methods model word alignment on a joint distribution, so that an incorrect previous alignment might cause more incorrect alignments like dominos.
Our experiments prove that their method consistently works for extended word alignment.


\subsection{Original Word Tag Prediction}
\label{sec:original tag prediction}
\begin{figure}
\centering
\includegraphics[width=75mm]{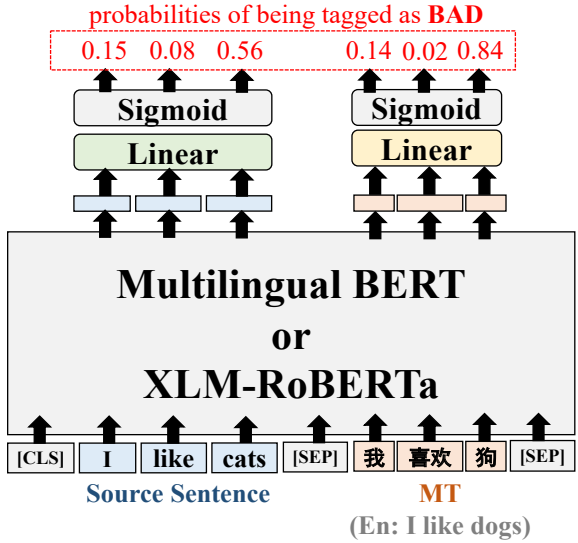}
\caption{
Determining original word tags with pre-trained language models.
}
\label{fig:seq_tag}
\end{figure}
For Original tags, we conduct sequence tagging with multilingual pre-trained language models including mBERT and XLM-R.
Figure \ref{fig:seq_tag} shows the image.
Input sequence is organized in the format of ``[CLS] \textit{source sentence} [SEP] \textit{MT} [SEP]''  without any mark tokens.
Two linear layers followed by Sigmoid function transform output vectors into scalar values as respective probabilities of being \textbf{BAD} for each token.
Formally, for a source sentence $S=[s_1,s_2,...,s_i, ...,s_m]$ and an MT $T=[t_1,t_2,...,t_j,...,t_n]$, the total loss is the mean of binary cross entropy of all word tags.
\begin{center}
$\mathcal{L}^{tag}_{s_i}=-[y_{s_i}\mathrm{log}(p_{s_i})+(1-y_{s_i})\mathrm{log}(1-p_{s_i})]$
$\mathcal{L}^{tag}_{t_j}=-[y_{t_j}\mathrm{log}(p_{t_j})+(1-y_{t_j})\mathrm{log}(1-p_{t_j})]$
$\mathcal{L}^{tag}=\frac{1}{m+n}(\sum\limits_{i=1}^m \mathcal{L}_{s_i} + \sum\limits_{j=1}^n \mathcal{L}_{t_j})$
\end{center}
We have also implemented our models with classification top-layers\footnote{Classification top-layers refers to a binary classification linear layer with Softmax.}, but we find that regression models are consistently better since we can adopt flexible threshold to offset the bias caused by imbalance of reference tags.

\subsection{Word Tag Refinement and Gap Tag Prediction}
\label{sec:word tag refinement}
\begin{figure}
\centering
\includegraphics[width=70mm]{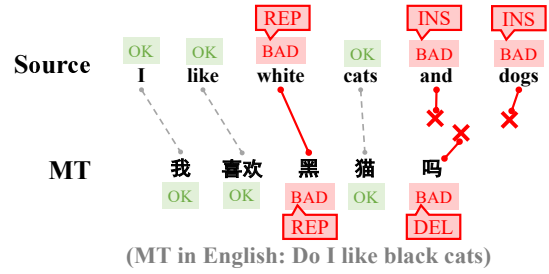}
\caption{Refining the word tags by using extended word alignment.}
\label{fig:tag by align}
\end{figure}
We use extended word alignment to refine the original word tags.
Following the rules described in Section \ref{subsec:refined_qe_tags}, we can refine word tags as Figure \ref{fig:tag by align} shows.
In practical situation, some \textbf{BAD}-tagged words are likely to be aligned with \textbf{OK}-tagged words.
In that case, we change \textbf{OK} into \textbf{BAD} encouraging more generation of \textbf{REP}.

\begin{figure}
\centering
\includegraphics[width=70mm]{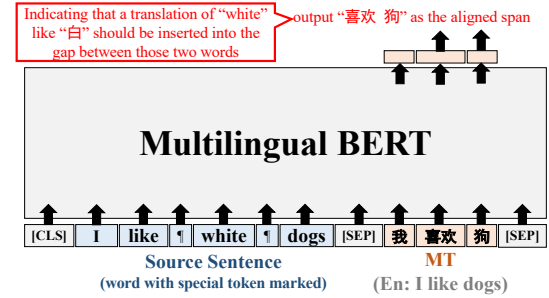}
\caption{
Determining gap tags by extracting source-gap alignments with mBERT.
}
\label{fig:src-gap align by mbert}
\end{figure}
For gap tags, we adopt a method similar to the one described in Section \ref{sec:align by mbert}.
Specifically, we model source-gap correspondences as alignment between source words and MT gaps.
We train a model that aligns an \textbf{INS}-tagged source word with a two-word span in MT where corresponding gap is surrounded.
Figure \ref{fig:src-gap align by mbert} illustrates our idea.
During testing, when a valid source-gap correspondence is confirmed, we tag the MT gap as \textbf{INS}\footnote{As for source words involved, we do not change their tags and trust the refinement based on extended word alignment because we believe extended word alignment is easier to model.}.

It is natural if we use such a method to determine gap tags based on the \textbf{INS}-tagged source words prediction from previous workflow.
However, in experiment, we notice that absolute value of accuracy of \textbf{INS}-tagged source words is not high.
In order not to be influenced by the previous wrong predictions, instead of treating this task as a downstream one, we conducted it independently.

\section{Experiment}
\begin{table*}[]
\small
\centering
\scalebox{0.95}[0.95]{
\begin{tabular}{ccc|cc}
\hline
& \multicolumn{2}{c|}{\textbf{En-De}}		& \multicolumn{2}{c}{\textbf{En-Zh}}  \\
& \textbf{Source MCC} & \textbf{MT MCC}		& \textbf{Source MCC}		& \textbf{MT MCC} \\ \hline \hline
\textbf{OpenKiwi}		& 0.266				& 0.358				& 0.248				& 0.520		\\ \hline
\textbf{mBERT-cls}		& 0.314				& 0.419				& 0.309				& 0.555		\\ \hline
\textbf{mBERT}			& 0.340				& \textbf{0.457}	& \textbf{0.357}	& 0.570		\\ \hline
\textbf{XLM-R-cls}		& 0.326				& 0.446				& 0.330				& 0.579		\\ \hline
\textbf{XLM-R}			& \textbf{0.345}	& 0.453				& 0.354				& \textbf{0.592}		\\ \hline \hline
\textbf{WMT20 Top}		& 0.523\citep{wang-EtAl:2020:WMT4}	& 0.597\citep{lee-2020-two}	& 0.336\citep{rubino-2020-WMT}	& 0.610\citep{hu-2020-WMT}	\\ \hline
\end{tabular}
}
\caption{
MCC of original tags. All MT gap tags of our systems are set to \textbf{OK}. For En-De, unlike top systems that employs large-scale third-party resources, we achieve acceptable performance only using QE dataset.
}
\label{table:original tag results}
\end{table*}
\begin{figure*}
\centering
\includegraphics[width=150mm]{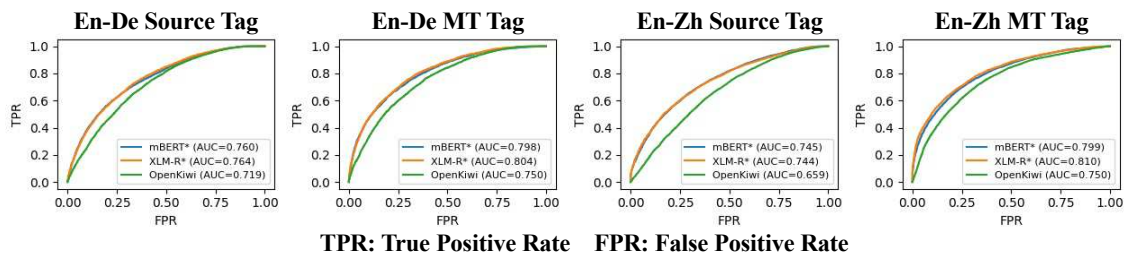}
\caption{ROC curve and AUC of the baseline and our systems(* indicates that the model outperforms the baseline (OpenKiwi) with statistical significance (p<0.01)).}
\label{fig:orig tag roc}
\end{figure*}
\subsection{Data and Experimental Setups}
We make full advantage of the En-De and En-Zh datasets of the shared task of original word-level QE in WMT20\footnote{\url{http://www.statmt.org/wmt20/quality-es}\\ \url{timation-task.html}}.
There are 7,000, 1,000, and 1,000 sentence pairs with annotation of tags respectively for training, development, and test set.
Since the original datasets do not contain refined objectives, we additionally annotate the original development sets with all the objectives for refined word-level QE.
Those annotated 1,000 pairs are further divided into 200 pairs for evaluation and 800 pairs for fine-tuning.
\begin{table*}[]
\small
\centering	
\begin{tabular}{c|c|c|c}
\hline
\textbf{Extended Word Align.} & \textbf{Source-Gap Corr.}       & \textbf{En-De (F1/P/R)} & \textbf{En-Zh (F1/P/R)} \\ \hline \hline
\textbf{FastAlign}            & \multirow{3}{*}{\textbf{mBERT}} & 0.828/0.812/0.844       & 0.739/0.773/0.709       \\ \cline{1-1} \cline{3-4} 
\textbf{AWESoME}  &                   & 0.891/0.915/0.868 & 0.814/0.871/0.764 \\ \cline{1-1} \cline{3-4} 
\textbf{mBERT}    &                   & 0.895/0.917/0.875 & 0.836/0.888/0.790 \\ \hline
\textbf{ft-mBERT} & \textbf{ft-mBERT} & \textbf{0.916}/0.913/0.918 & \textbf{0.888}/0.887/0.889 \\ \hline
\end{tabular}
\caption{
Evaluation of word-level correspondences.
``mBERT'' indicates mBERT trained with 7,000-pair pseudo data and ``ft-mBERT'' indicates mBERT further fine-tuned with 800-pair data.
}
\label{table:align results}
\end{table*}

All the experiments are conducted with modified scripts from transformers-v3.3.1\footnote{\url{https://github.com/huggingface/transfo} \\ \url{rmers}} on an NVIDIA TITAN RTX (24GB) with CUDA 10.1.
For pre-trained models, we use \textit{bert-base-multilingual-cased} for mBERT and \textit{xlm-roberta-large} for XLM-R from Huggingface.

To train the model for original tags described in Section \ref{sec:original tag prediction}, we use the 7,000-pair training set provided by WMT20.
800 pairs of manually annotated data whose refined tags are degenerated into original tags are used for further training.
Learning rate is set to 3e-5 and 1e-5 for mBERT and XLM-R respectively and both models are trained for 5 epochs.
All the other configurations are remained unchanged as the default.

To train the models extracting extended word alignment described in Section \ref{sec:align by mbert}, we utilize AWESoME \citep{dou-2021-word} to generate pseudo alignment data based on 7,000-pair WMT20 training set.
We also use extra 800 sentence-pair annotated alignment data for fine-tuning.
Models are pre-trained for 2 epochs and fine-tuned for 5 epochs with a learning rate of 3e-5.
Most configurations are remained unchanged as the default but \textit{max\_seq\_length} and \textit{max\_ans\_length} are set to 160 and 15 following \citet{nagata-etal-2020-supervised}.

To train the model extracting source-gap correspondences described in Section \ref{sec:word tag refinement}, similar to what described above, we firstly adopt 7,000 sentence-pair WMT20 training set, generating pseudo data by randomly dropping out some target words in PE\footnote{Provided [P]ost-[E]dited sentence from MT in WMT20 dataset. It is regarded as the correct translations.}.
Then we link gaps where words are dropped with their source counterparts according to the source-PE alignment extracted by AWESoME.
Also, 800 sentence-pair of gold source-gap correspondences are used for fine-tuning.
All model configurations and training settings are kept identical as those of the model for extended word alignment extraction.


\subsection{Experimental Results}
\label{subsec:results}

\subsubsection{Evaluation of Original Tags}
\label{subsubsec:original qe tags results}
We firstly compare our performance with other participants of WMT20
Therefore, we use identical test sets to evaluate and only use data from the original training set of WMT20 to train our models here.
Following WMT20 \citep{WMT20}, we adopt the Matthews correlation coefficient (MCC) as the metric.
From the perspective of competition, we make every effort to boost the performance.
Thus we set all gap tags as \textbf{OK} rather than predicting them as we find such a strategy leads to the best MCC.
The results are shown in Table \ref{table:original tag results}.

In general, pre-trained language models consistently outperform the baseline which is a LSTM-based predictor-estimator implemented with OpenKiwi.
For En-De, our best source and MT MCC would have ranked sixth on the leaderboard of WMT20.
For En-Zh, our best source and MT MCC would have ranked first and second on the leaderboard of WMT20.

It is also noteworthy that regression models consistently outperform classification models with the suffix ``-cls''.
For regression models, we search an optimized threshold that maximize sum of source and MT MCC on the development set and adopt it on the test set to determine tags.
To exclude errors caused by single optimized threshold, we further draw the ROC curves and AUC in Figure \ref{fig:orig tag roc}.
The results demonstrate that our regression models based on mBERT and XLM-R statistically significantly outperform the baseline.

For En-De, \citet{wang-EtAl:2020:WMT4} and \citet{lee-2020-two} both used large-scale third-party data\footnote{\citet{wang-EtAl:2020:WMT4} used parallel data from WMT20 news translation task to pre-train a predictor and \citet{lee-2020-two} generated ~11 million pairs of pseudo QE data with ~23 million pairs of sentences.}.
Besides the top-2, the third system \citep{rubino-2020-WMT} is also pre-trained with 5 million sentence pairs but got 0.357 and 0.485 respectively.
Therefore, we believe that we achieve acceptable performance with very small cost.

\subsubsection{Evaluation of Word-Level Correspondences}
\label{subsubsec:extended word alignment results}
We evaluate extended word alignment and source-gap correspondences jointly as word-level correspondences.
The results are shown in Table \ref{table:align results}.
Two baselines (``FastAlign'' and ``AWESoME'') cannot predict source-gap correspondences since they are designed for ordinary word alignment.
We combine their extended word alignment with prediction of source-gap correspondences by ``mBERT'' for fair comparison.
All predictions are evaluated by F1 score as well as precision and recall.

Neural-based methods significantly outperform statistical ``FastAlign''.
The gap of 0.4\% for En-De and 2.2\% for En-Zh between ``AWESoME'' and ``mBERT'' is not significant.
But it might implies that pre-trained language models like mBERT is able to filter noises in pseudo data and produce high-quality word-level correspondences.
Additionally, a better performance of ``fine-tuned mBERT'' indicates that the upper bound could be higher if more annotated data is available.

\subsubsection{Evaluation of Refined Tags}
\label{subsubsec:refined qe word tags results}
\begin{table*}[]
\small
\centering
\subtable[En-De Results]{
\begin{tabular}{c|c|l|l}
\hline	
\textbf{\begin{tabular}[c]{@{}c@{}}Extended\\ Word Alignment\end{tabular}} &
  \textbf{\begin{tabular}[c]{@{}c@{}}Original\\ QE Tags\end{tabular}} &
  \textbf{\begin{tabular}[c]{@{}c@{}}Source F1 Scores\\ Mean (OK/REP/INS)\end{tabular}} &
  \textbf{\begin{tabular}[c]{@{}c@{}}MT F1 Scores\\ Mean (OK/REP/DEL/INS)\end{tabular}} \\ \hline \hline
\textbf{FastAlign}                 & \multirow{2}{*}{\textbf{OpenKiwi}} & 0.626 (0.696/0.492/0.174) & 0.767 (0.847/0.477/0.124/0.156) \\ \cline{1-1} \cline{3-4} 
\textbf{AWESoME}                   &                                    & 0.708 (0.781/0.549/0.373) & 0.807 (0.879/0.548/0.395/0.156) \\ \hline
\multirow{2}{*}{\textbf{mBERT}}    & \textbf{mBERT}                     & 0.739 (0.825/0.540/0.421) & 0.820 (0.895/0.544/0.389/0.156) \\ \cline{2-4} 
                                   & \textbf{XLM-R}                     & 0.709 (0.781/0.548/0.410) & 0.809 (0.879/0.522/0.415/0.156) \\ \hline
\multirow{2}{*}{\textbf{ft-mBERT}} & \textbf{rt-mBERT}                  & \textbf{0.755} (0.850/0.538/0.400) & \textbf{0.827} (0.904/0.535/0.347/0.175) \\ \cline{2-4} 
                                   & \textbf{rt-XLM-R}                  & 0.685 (0.748/0.544/0.431) & 0.805 (0.871/0.538/0.580/0.175) \\ \hline
\end{tabular}
}

\subtable[En-Zh Results]{
\begin{tabular}{c|c|c|c}
\hline
\textbf{\begin{tabular}[c]{@{}c@{}}Extended\\ Word Alignment\end{tabular}} &
  \textbf{\begin{tabular}[c]{@{}c@{}}Original\\ QE Tags\end{tabular}} &
  \textbf{\begin{tabular}[c]{@{}c@{}}Mean Source F1 Scores\\ (OK/REP/INS)\end{tabular}} &
  \textbf{\begin{tabular}[c]{@{}c@{}}Mean MT F1 Scores\\ (OK/REP/DEL/INS)\end{tabular}} \\ \hline \hline
\textbf{FastAlign}                 & \multirow{2}{*}{\textbf{OpenKiwi}} & 0.360 (0.379/0.280/0.071) & 0.728 (0.781/0.276/0.173/0.042) \\ \cline{1-1} \cline{3-4} 
\textbf{AWESoME}                   &                                    & 0.371 (0.391/0.285/0.066) & 0.733 (0.786/0.280/0.202/0.042) \\ \hline
\multirow{2}{*}{\textbf{mBERT}}    & \textbf{mBERT}                     & 0.836 (0.914/0.446/0.020) & 0.891 (0.947/0.441/0.316/0.042) \\ \cline{2-4} 
                                   & \textbf{XLM-R}                     & 0.843 (0.929/0.410/0.018) & 0.895 (0.955/0.402/0.275/0.042) \\ \hline
\multirow{2}{*}{\textbf{ft-mBERT}} & \textbf{rt-mBERT}                  & 0.848 (0.929/0.447/0.034) & 0.897 (0.954/0.441/0.284/0.042) \\ \cline{2-4} 
                                   & \textbf{rt-XLM-R}                  & \textbf{0.849} (0.928/0.451/0.028) & \textbf{0.897} (0.955/0.446/0.289/0.042) \\ \hline
\end{tabular}
}
\caption{
Evaluation of refined tags.
Main metric is a weighted mean of F1 scores according to ratio of each type of tags in reference.
``ft-'' indicates that the model is fine-tuned with the extra 800-pair annotated alignment data.
``rt-'' indicates that the model is further trained with extra 800-pair annotated tag data.
}
\label{table:refined results}
\end{table*}
As introduced, we combine prediction of extended word alignment and original word tags\footnote{While predicting the original tags, we did not directly use the optimized threshold determined in Section \ref{subsubsec:original qe tags results} since test set here originates from the original development set. Instead, we take the original test set of WMT20 for development purposes and re-searched an optimized threshold on it.} to get refined word tags.
Moreover, we deduce gap tags from source-gap correspondences.
Origin of source-gap correspondences used is kept consistent with Table \ref{table:align results} according to extended word alignment.
For baseline, combinations of FastAlign/AWESoME and OpenKiwi is adopted.
As for metric, we use F1 score of each type of tags along with a weighted mean of all those F1 scores, taking the proportion of each tag in reference as weight.
The results are shown in Table \ref{table:refined results}.

Our best model outperforms the baseline by 12.9\% and 6.0\% respectively on source and MT refined tags in terms of mean F1 scores in En-De experiments.
As for the En-Zh experiments, mean F1 scores are significantly improved by 48.9\% and 16.9\%.

We also notice that though fine-tuned mBERT extracts extended word alignment with good accuracy, the absolute value of refined tag accuracy is still unsatisfactory (especially that of \textbf{INS} and \textbf{DEL}).
We will discuss that in the next section.

\section{Discussion on Specific Cases}
\label{sec:case study}
\begin{figure*}
\centering
\subfigure[An En-Zh case with correct refined word tag prediction.]{
\includegraphics[width=75mm]{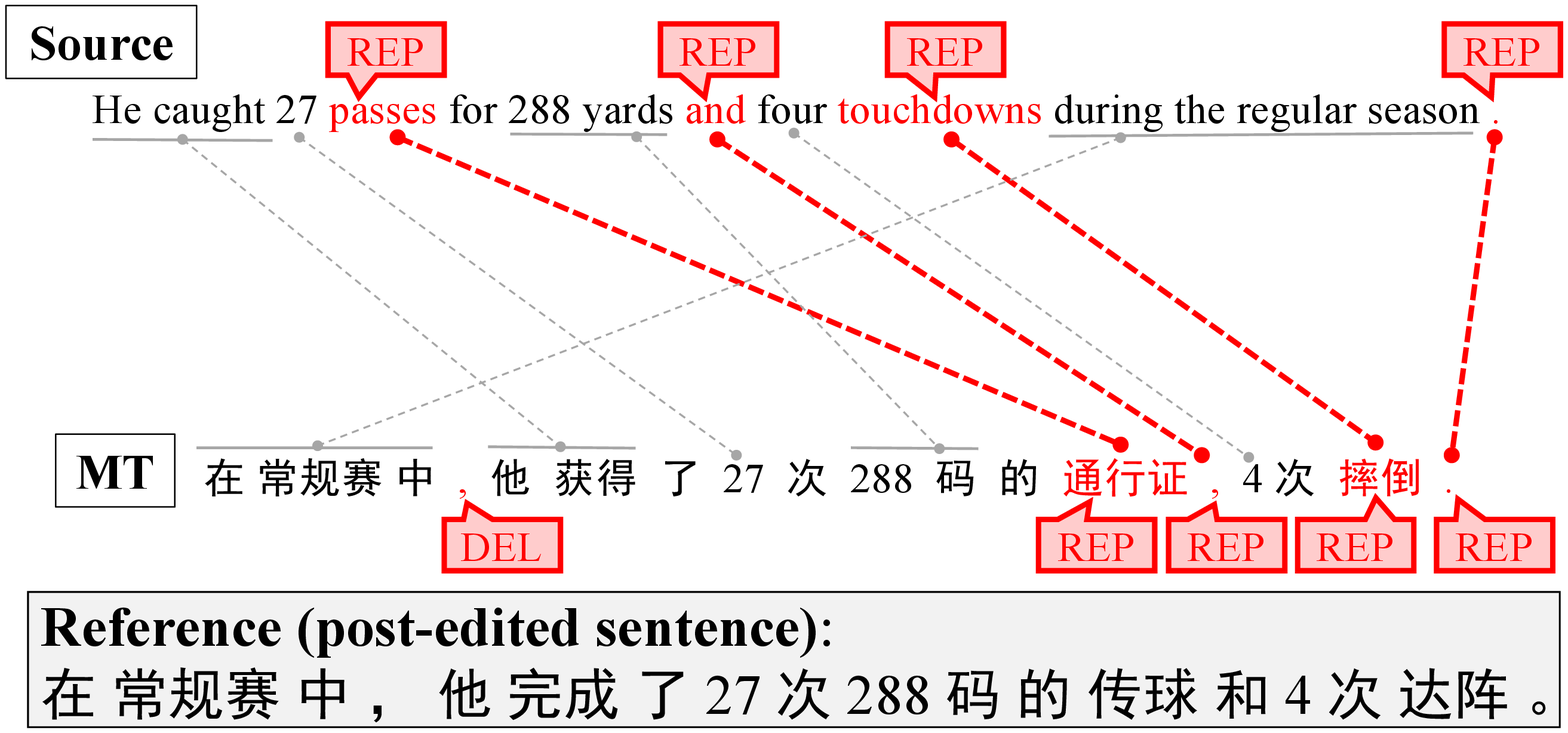}
\label{fig:case_good}
}
\subfigure[An En-Zh case with incorrect refined word tag prediction.]{
\includegraphics[width=75mm]{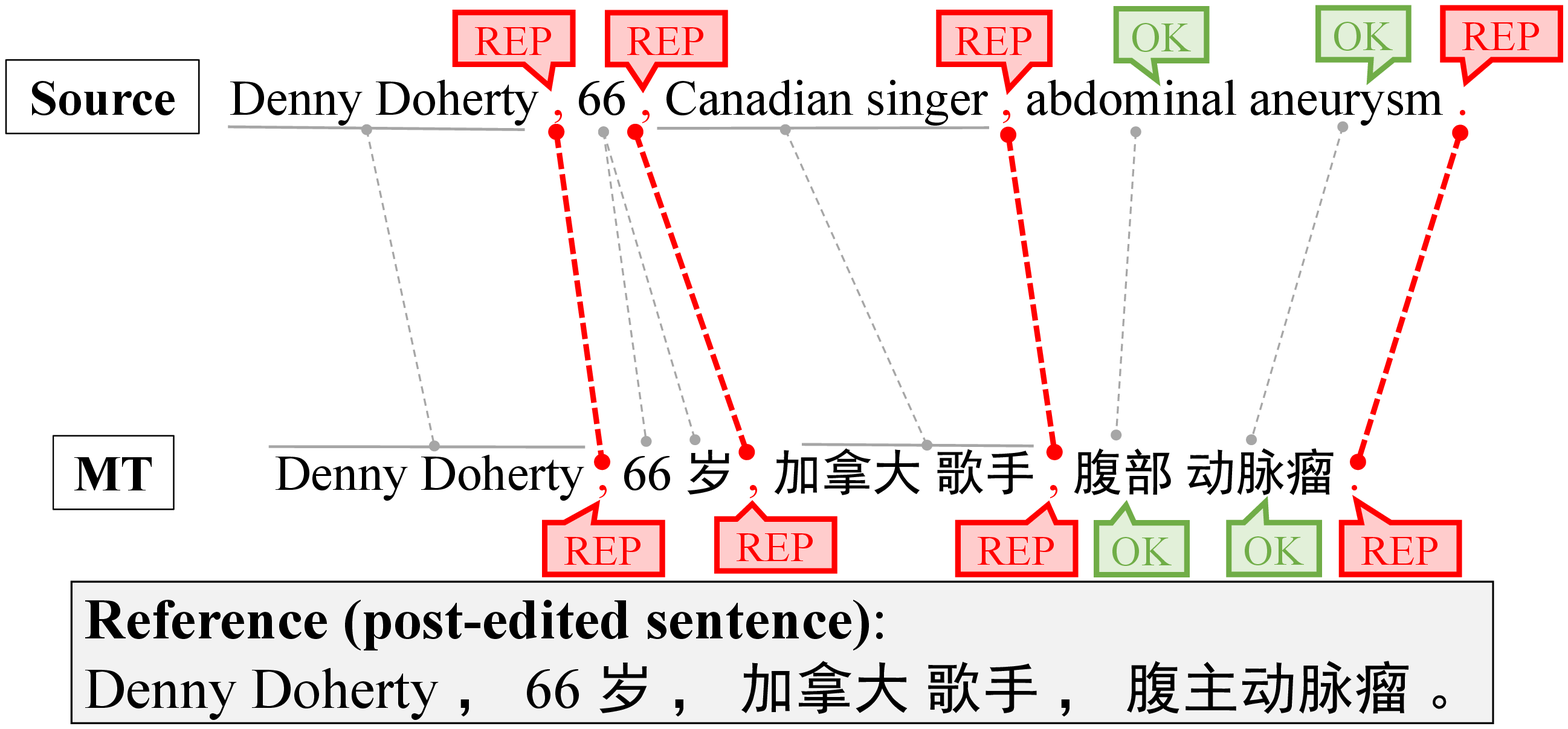}
\label{fig:case_bad}
}
\subfigure[An En-De case with correct prediction of source-gap correspondence and gap tag.]{
\includegraphics[width=75mm]{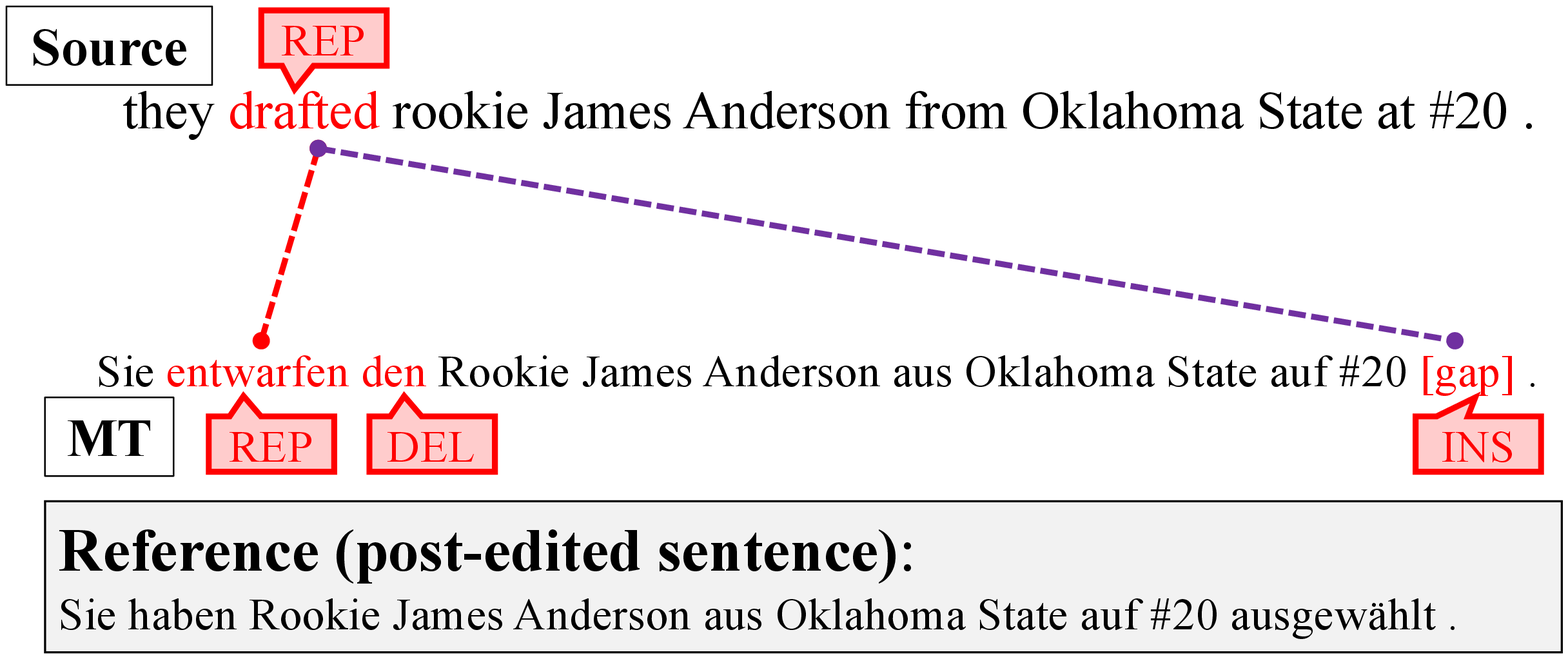}
\label{fig:case_ende_good}
}
\subfigure[An En-Zh case with incorrect prediction of source-gap correspondences and gap tags.]{
\includegraphics[width=75mm]{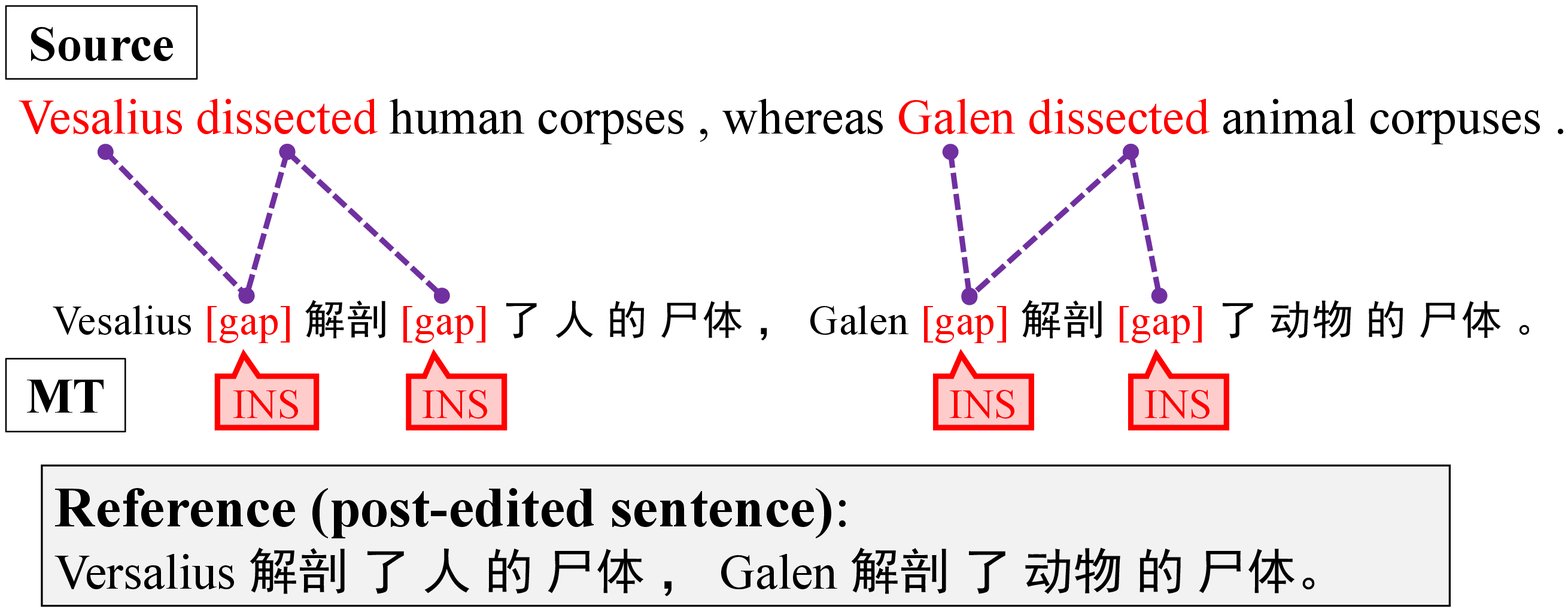}
\label{fig:case_enzh_bad}
}
\caption{Specific cases. For visual neatness, most \textbf{OK} tags are omitted and some continuous spans are merged.}
\label{fig:case}
\end{figure*}
\subsection{Discussion on Refined Word Tags}

In Figure \ref{fig:case_good}, our system basically succeeds in detecting errors caused by incorrect use of punctuations.
Our system correctly suggests the replacements for the second comma and the half-width period.
As for the first comma, the translation is still natural and acceptable if we delete the comma following the system's suggestion.
Moreover, our system successfully detects mistranslations of ``passes'' and ``touchdowns''.
In MT, those football terminologies are respectively translated as ``\begin{CJK}{UTF8}{gbsn}通行证\end{CJK}'' (a pass to enter somewhere) and ``\begin{CJK}{UTF8}{gbsn}摔倒\end{CJK}'' (falling down).
It is noteworthy that those two mistranslations are not revised in post-edited corpus provided by WMT21.
It implies that our system performs surprisingly well as it even succeeds in detecting mistranslations that is not noticed by human annotators.

In Figure \ref{fig:case_bad}, our system still works well in detecting incorrect use of half-width punctuations.
However, compared with the reference, ``abdominal aneurysm'' is mistranslated and our model failed to detect it because both are tagged as \textbf{OK} during the prediction of original tags.
A premature prediction of \textbf{OK} prevents a word from being refined into \textbf{REP}/\textbf{INS}/\textbf{DEL} later.

We believe that an inappropriate threshold mainly leads to such an issue.
Predicted probabilities of ``\begin{CJK}{UTF8}{gbsn}腹部\end{CJK}'' and ``\begin{CJK}{UTF8}{gbsn}动脉瘤\end{CJK}'' are respectively 0.103 and 0.134, but the optimized threshold used is 0.88 as we searched it to maximize the MCC on the whole set.
Meanwhile, probabilities of all other \textbf{OK}-tagged MT words are actually smaller than 0.01.
As a result, if we set a threshold between 0.01-0.10 for this sentence pair, we could have obtained the perfect result.
In the future, we plan to investigate into methods that can determine fine-grained optimized threshold for each sentence pair.

\subsection{Discussion on Gap Tags}

Figure \ref{fig:case_ende_good} shows a typical En-De case that our model handles well.
In German, it is more natural to indicate actions took place in the past in perfect tense rather than past tense.
In this case, English verb ``drafted'' should be modified to ``haben ... ausgewählt''.
Our model correctly suggests a correspondence between ``drafted'' and the MT gap in front of the period.
As there are many cases need similar modification that inserts a particular word (like particle or infinitive for clause) before the period in MT, it is easier for our model to learn such laws.
It probably explains the relatively good accuracy of \textbf{INS} in En-De experiments.

In contrast, Figure \ref{fig:case_enzh_bad} is an En-Zh example showing that our model tends to align many source words with the gap right before or after their translation in MT even the translation is correct and needs no extra insertions.
The word ``dissected'' is unnecessarily aligned with the gap around its translation ``\begin{CJK}{UTF8}{gbsn}解剖\end{CJK}''.
Two human names are also unnecessarily aligned with gaps.
As a result, four gaps are incorrectly tagged as \textbf{INS}.
We observed the annotated dataset and noticed that many Chinese words in MT are slightly modified by adding prefixes and suffixes during post-editing.
For example, ``\begin{CJK}{UTF8}{gbsn}成年 海龟\end{CJK}'' (adult sea turtle) is modified to ``\begin{CJK}{UTF8}{gbsn}成年的海龟\end{CJK}'' (adding ``\begin{CJK}{UTF8}{gbsn}的\end{CJK}'' as a suffix for adjective).
``\begin{CJK}{UTF8}{gbsn}演讲\end{CJK}'' (the speech) is modified to ``\begin{CJK}{UTF8}{gbsn}这一演讲\end{CJK}'' (emphasizing ``this'' speech).
Generally, those modifications are not necessary because of the free Chinese grammar.
However, existence of those modifications might mislead the model into preferring to unnecessarily align a word with the gap around its translation like Figure \ref{fig:case_enzh_bad}.
To address this issue, we plan to restrict the annotation rules to exclude meaningless modification in En-Zh training data in the future.

\section{Conclusion and Future Work}
To improve post-editing assistance efficiency, we define a novel concept called extended word alignment.
By incorporating extended word alignment with original word-level QE, we formally propose a novel task called refined word-level QE.
To solve the task, we firstly adopt a supervised method to extract extended word alignment and then predict original tags with pre-trained language models by conducting sequence tagging.
We then refine word tags with extended word alignment.
Additionally, we extract source-gap correspondences and determine gap tags.
We perform experiments and a discussion on specific cases.

In the future, we would like to polish our work in the following perspectives.
Firstly, we want to develop methods that determines fine-grained threshold as elaborated in Section \ref{sec:case study}.
Moreover, we prepare to conduct a human-evaluated experiment to prove the superiority of refined word-level QE in terms of post-editing assistance efficiency.

\bibliographystyle{acl_natbib}
\bibliography{main}

%

\end{document}